%% file: main.tex
\documentclass[sigconf]{acmart}
\input{commands}

\AtBeginDocument{%
  }

\copyrightyear{2024}
\acmYear{2024}
\setcopyright{rightsretained}
\acmConference[ESEM '24]{Proceedings of the 18th ACM / IEEE International Symposium on Empirical Software Engineering and Measurement}{October 24--25, 2024}{Barcelona, Spain}
\acmBooktitle{Proceedings of the 18th ACM / IEEE International Symposium on Empirical Software Engineering and Measurement (ESEM '24), October 24--25, 2024, Barcelona, Spain}

\begin{document}

\title{Exploring LLM-Driven Explanations for Quantum Algorithms}

\author{Giordano d'Aloisio}
\orcid{0000-0001-7388-890X}
\affiliation{%
    \institution{University of L'Aquila}
    \city{L'Aquila}
    \country{Italy}
}
\email{giordano.daloisio@graduate.univaq.it}

\author{Sophie Fortz}
\orcid{0000-0001-9687-8587}
\affiliation{%
    \institution{King's College London}
    \city{London}
    \country{United Kingdom}
}
\email{sophie.fortz@kcl.ac.uk}

\author{Carol Hanna}
\orcid{0009-0009-7386-1622}
\affiliation{%
    \institution{University College London}
    \city{London}
    \country{United Kingdom}
}
\email{carol.hanna.21@ucl.ac.uk}

\author{Daniel Fortunato}
\orcid{0000-0003-2596-6859}
\affiliation{%
    \institution{University of Porto}
    \city{Porto}
    \country{Portugal}
}
\email{dabf@fe.up.pt}

\author{Avner Bensoussan}
\orcid{0009-0007-3285-9468}
\affiliation{%
    \institution{King's College London}
    \city{London}
    \country{United Kingdom}
}
\email{avner.bensoussan@kcl.ac.uk}

\author{Eñaut Mendiluze Usandizaga}
\orcid{0009-0007-3315-1664}
\affiliation{%
    \institution{Simula Research Laboratory}
    \city{Oslo}
    \country{Norway}
}
\email{enaut@simula.no}

\author{Federica Sarro}
\orcid{0000-0002-9146-442X}
\affiliation{%
    \institution{University College London}
    \city{London}
    \country{United Kingdom}
}
\email{f.sarro@ucl.ac.uk}


\renewcommand{\shortauthors}{d'Aloisio et al.}

\begin{abstract}
\textbf{Background:} Quantum computing is a rapidly growing new programming paradigm that brings significant changes to the design and implementation of algorithms.  Understanding quantum algorithms requires knowledge of physics and mathematics, which can be challenging for software developers. \textbf{Aims:} In this work, we provide a first analysis of how LLMs can support developers' understanding of quantum code. \textbf{Method:} We empirically analyse and compare the quality of explanations provided by three widely adopted LLMs (Gpt3.5, Llama2, and Tinyllama) using two different human-written prompt styles for seven state-of-the-art quantum algorithms. We also analyse how consistent LLM explanations are over multiple rounds and how LLMs can improve existing descriptions of quantum algorithms. \textbf{Results:} Llama2 provides the highest quality explanations from scratch, while Gpt3.5 emerged as the LLM best suited to improve existing explanations. In addition, we show that adding a small amount of context to the prompt significantly improves the quality of explanations. Finally, we observe how explanations are qualitatively and syntactically consistent over multiple rounds. \textbf{Conclusions:} This work 
highlights promising results, and opens challenges for future research in the field of LLMs for quantum code explanation. Future work includes refining the methods through prompt optimisation and parsing of quantum code explanations, as well as carrying out a systematic assessment of the quality of explanations.
\end{abstract}

\ccsdesc[300]{Computer systems organization~Quantum computing}
\ccsdesc[300]{Computing methodologies~Artificial intelligence}

\keywords{Quantum Computing, Large Language Models, Code Explainability.}


\maketitle              

\section{Introduction}\label{sec:intro}
Quantum computing is a multi-disciplinary field comprising computer science, physics, and mathematics, which utilises quantum mechanics to solve specific problems much faster than a classical computer would do~\cite{grover,shor}.
Designing and implementing quantum algorithms fundamentally differs from doing so for classical algorithms. Developing quantum algorithms requires skills in mathematics and physics that software developers lack, which creates an entry barrier for most developers interested in quantum computing. 

Lowering this entry barrier should be a priority for the continuously growing quantum software engineering community~\cite{Zhao_2021}. 
We hypothesise that using large language models (LLMs) to generate high-quality explanations of quantum programs can lower the entry barrier for researchers or developers interested in quantum software engineering. Thus, in this work, we investigate LLMs's ability to automatically generate explanations for quantum programs. 
Specifically, we carry out an empirical study to investigate the ability of three widely adopted LLMs (i.e., Gpt3.5, Llama2, and Tinyllama) to generate explanations for seven state-of-the-art quantum algorithms written in the OpenQSAM programming language~\cite{cross2018ibm} when no context is given to the LLMs. In addition, we investigate whether adding a small amount of context to the prompt improves the quality of explanations, and how consistent the generated explanations are over multiple rounds of generation. Finally, we analyse the ability of LLMs to improve existing descriptions of quantum algorithms.  
All the explanations generated were manually rated by four human-raters with different expertise in software engineering and different degree of familiarity with quantum computing.

According to the human-raters, Llama2 is the LLM providing the best explanations when no context is provided, while Gpt3.5 is better suited to improve already existing explanations.
Furthermore, we observe that providing the LLMs with a small amount of context (i.e., just the name of the algorithm implemented and the number of qubits employed) significantly improves the quality of explanations for all the LLMs analysed herein. Finally, we found no difference (both qualitative and syntactical) in explanations generated over multiple rounds, thus proving a good level of stability of the LLMs towards this task. 

The promising results obtained in this preliminary analysis encourage further research on the use of LLMs' explanations of quantum code, including exploring prompt optimisation and conducting a larger human-based evaluation.

The contributions of our work are as follows:
\begin{itemize}
    \item To the best of our knowledge, this is the first work providing insights on LLMs' ability to explain quantum code;
    \item We detail future research directions in assessing and improving quantum code explanations;
    \item We provide the full replication package of our work, including the quantum algorithms employed and the scripts to generate explanations \cite{giordano_d_aloisio_2024_12805237}.
\end{itemize}


\section{Background and Motivation}\label{sec:background}

Quantum programming is an emergent discipline born from recent advancements in quantum mechanics. The era of quantum computing is still in its infancy, marked by the development of the first quantum computers. Despite its early stage, software engineers are increasingly recognising the potential of quantum computing, as evidenced by the surge in quantum-related papers presented at recent software engineering conferences~\cite{barbosa2020software, de2022software, murillo2024challenges, ali2022software, agrawal2023quantum, miranskyy2021testing, di2024need, leite2024testing, li2021understanding}.  However, the high entry barrier remains a significant obstacle to popularising the field.

Fundamentally, a quantum program is a circuit where each wire represents a bit or a qubit and employs two types of operators: quantum gates and measurement operators. Quantum gates, usually defined as matrices, perform operations on one or more qubits, altering their states. Measurement operators, on the other hand, extract the result of the quantum circuit and transform it into classical bits that can be interpreted by conventional computers.

Physicists have provided tools to manipulate quantum information, such as languages like OpenQASM. From a physicist's perspective, these tools fulfil the requirement, as software engineers no longer need to manipulate physical qubits directly but can use abstract representations consisting of logical qubits and gates. However, from a software engineer's viewpoint, these languages may seem overly complex, resembling writing code solely with logical circuits.

Despite these advancements, the field of quantum programming still faces significant challenges, particularly in making the technology accessible to a broader range of software engineers. Bridging this gap requires not only the development of more intuitive programming languages and tools but also employing comprehensive resources to lower the entry barrier and foster wider adoption and innovation in quantum computing.

This paper centres on the latter idea: employing resources to diminish the entry barrier and catalyse wider adoption and innovation in quantum computing. Since the release of ChatGPT in November 2022, large language models (LLMs) have been assisting developers and software engineers in several tasks. Among those, code comprehension emerged as one of the tasks in which LLMs have been mostly involved \cite{fan2023large}. Hence, we believe that LLMs can be a useful resource to help developers and software engineers understand quantum code. However, as mentioned above, quantum code can not be considered at the same level as traditional code, still being more assembly-level style. Therefore, systematically assessing the ability of LLMs to provide specific explanations for quantum algorithms is still a challenge that has to be addressed. This paper provides a first step in this direction: we engage three LLMs to provide code explanations for seven quantum algorithms using prompts providing different amounts of context and ask human raters to assess each explanation for correctness and potential comprehension enhancement.

\section{Related Work}\label{sec:related}
LLMs have been used in previous work to explain different aspects of code. For instance, Sarsa \emph{et al.}~\cite{sarsa2022automatic} employ LLMs to generate code explanations for educational purposes. 
Also focusing on education, MacNeil \emph{et al.}~\cite{macneil2022generating} analyse how LLMs are able to explain numerous aspects of a given code snippet and Sobania \emph{et al.}~\cite{sobania2023evaluating} empirically analyse the ability of LLMs to explain software patches. Balfroid \emph{et al.}~\cite{balfroid2024towards} use ChatGPT to provide code tours for code onboarding \emph{i.e.,} the process of transitioning new employees into a team's methodology and tools. Leinonen \emph{et al.}~\cite{leinonen2023using} use LLMs to improve programming error messages. Concerning the integration between LLMs and quantum code, Easttom~\cite{easttom2024utilizing} analyse the ability of ChatGPT to generate and improve quantum algorithms. Guo \emph{et al.}~\cite{guo2024repairing} propose to use ChatGPT for automated quantum program repair, while Ezratty highlights how LLMs can be employed in different aspects of Quantum Computing, such as learning, software development or research \cite{olivier2023}. 
However, to the best of our knowledge, no work has yet attempted to analyse the ability of LLMs to explain quantum code. Our work seeks to fill this gap providing first results and shepherding the way for future research in this field.

\section{Methodology}
\label{sec:method}

In this section, we describe the methodology followed for our evaluation. 
We aim to answer the following research questions (\textbf{RQs}):

\noindent\ding{228} \textbf{RQ$_1$:} \emph{Do different LLMs generate quantum algorithm explanation of different quality?} This RQ focuses on assessing the overall quality of the explanations generated by the LLMs and whether there is a model that is more suited for explaining quantum code.

\noindent\ding{228} \textbf{RQ$_2$:} \emph{To what extend does adding context to the prompt impact the explanation quality?} This RQ investigates whether providing an LLM with a prompt including some context (in our case, only the name of the algorithm and the number of qubits are provided) improves the overall quality of explanations.

\noindent\ding{228} \textbf{RQ$_3$:} \emph{How consistent are the LLM explanations over different runs?} Given the stochastic nature of LLMs \cite{saba2023stochastic}, this RQ aims to assess if running a prompt multiple times leads to both qualitatively and syntactically similar explanations.

\noindent\ding{228} \textbf{RQ$_4$:} \emph{Do different LLMs exhibit a different ability in improving existing explanations for quantum code?} Instead of explaining a quantum algorithm from scratch, this RQ focuses on assessing how able an LLM is to improve existing descriptions.

In the following, we first present the quantum algorithms employed in our evaluation. Then, we describe the process followed to generate explanations from the LLMs. Finally, we report our the evaluation process we followed to answer each RQ.

\subsection{Evaluation Benchmark} 

For our evaluation, we employ seven quantum algorithms selected from the MqtBench~\cite{mqtbench} benchmark written in the OpenQASM 3 programming language \cite{qasm3} for the Qiskit quantum compiler. The selected algorithms are the following: \emph{Amplitude Estimation (AE)}: this algorithm aims to find an estimation for the amplitude of a certain quantum state \cite{suzuki2020amplitude}; \emph{Deutsch-Jozsa (DJ)}: this algorithm determines, whether an unknown oracle mapping input values either to 0 or 1 is constant (always output 1 or always 0) or balanced (both outputs are equally likely) \cite{collins1998deutsch}; \emph{Grover}: Grover's algorithm finds a certain goal quantum state determined by an oracle \cite{long2001grover}; \emph{Quantum Fourier Transform (QFT)}: this algorithm embodies the quantum equivalent of the discrete Fourier transformation \cite{weinstein2001implementation}; \emph{Quantum Fourier Transform with entanglement (QFT-ent)}: this algorithm is a variation of QFT to entangled qubits; \emph{Quantum Phase Estimation (QPE)}: this algorithm estimates the phase of a quantum operation \cite{dorner2009optimal}; \emph{Quantum Walk (QW)}: the quantum equivalent to classical random walks \cite{venegas2012quantum}. The rationale for the selection of these algorithms is twofold; they are among the most popular quantum algorithms, and they are used in other empirical studies~\cite{Fortunato_Campos_Abreu_2022}. Moreover, we adopted the OpenQASM implementation of these algorithms instead of their Python ones to better assess the ability of LLMs to explain algorithms written in less common programming languages.

\subsection{Generation of Code Explanations}
We analyse explanations generated by using three different LLMs, namely: Gpt3.5 turbo 16k, Llama2, and Tinyllama. We have chosen Gpt3.5 and Llama2 because they are two of the most widely adopted general-purpose LLMs. Moreover, we also analyse explanations generated from Tinyllama to check if a smaller LLM is also able to generate good-quality quantum code explanations. Concerning Llama and Tinyllama, we adopt the implementations provided by the \texttt{ollama} ecosystem with their default hyper-parameters,\footnote{\url{https://ollama.com/}} while for Gpt3.5 we employ the model provided by the OpenAI API\footnote{\url{https://openai.com/index/openai-api/}} and, following a previous work about quality assessment of code explanations from Gpt3.5 \cite{sobania2023evaluating}, we set its \texttt{temperature} to $0.8$.

To generate the code explanations for \textbf{RQ$_1$, RQ$_2$, and RQ$_3$}, we feed each LLM with two different prompt styles: the first style, which we call \textit{Non Context-Aware}, does not provide any information about the quantum code we feed as input: 

\textit{"Can you give a high-level explanation of this code?} $<$algorithm code$>$"

The second style, which we call \textit{Context-Aware}, includes a few basic information about the quantum code feed as input, i.e., the name of the algorithm implemented by the quantum code and the number of qubits (this information was obtained from the MqtBench benchmark): 

\textit{"Can you give a high-level explanation of this code?} $<$algorithm code$>$. 
\textit{The name of the algorithm is:} $<$algorithm name$>$. 
\textit{The code includes} $<$number of qubits$>$ \textit{qubits"}

The structure of the above prompts has been extensively discussed and agreed by all the authors of the paper. 
Note that we ask the LLM to provide \emph{high-level} explanations for both prompts to assess better the LLM's ability to understand an algorithm's behaviour instead of just describing each line of code.

Finally, to address the consistency of the explanations, following the methodology employed in previous work \cite{sobania2023evaluating}, we repeated the generation process three times for each LLM (3), quantum algorithm (7), and prompt style (2) combination, yielding a total of 126 different explanations. 

Concerning \textbf{RQ$_4$}, we extract from the MqtBench repository a short description for each quantum algorithm and ask the LLMs to improve it using the following prompt:

\textit{"Can you improve the following explanation:} $<$code explanation$>$
\textit{for the following quantum algorithm:} $<$algorithm code$>$
\textit{named}: $<$algorithm name$>$
\textit{making it more informative but also keeping it simple?"}

Like the previous case, this prompt has been discussed and approved by all the authors. In this context, however, we perform only one round of generation due to limited computational resources and human-effort needed to rate multiple explanations. Future works can investigate the consistency of description improvements over multiple rounds of generation. 


\subsection{Evaluation Process and Metrics}\label{sec:eval}

After collecting the different explanations, four independent respondents evaluate them. These are software engineers (SE) working in quantum computing. We employed SE with knowledge of quantum computing because they can provide a more reliable evaluation of quantum explanations and are more able to detect errors in the explanations compared with people with no expertise on the topic. 
In particular, two of the evaluators reported 3-4 years of experience in SE, while the other two have 5-10 years. As for the experience in quantum computing, one respondent reported less than a year of experience, two have 1-2 years, and one has 3-4 years, highlighting a heterogeneous group of raters.

We asked them to give a score from $1$ to $5$ for each explanation for each quantum algorithm, where $1$ means an entirely wrong explanation, while $5$ indicates a high-quality explanation. To help the evaluators assess the correctness and quality of each explanation, we provided them with the source code of each algorithm and the one-line description of the algorithm, as given by MqtBench~\cite{mqtbench}.  

To avoid potential bias during the human evaluation, we anonymise the LLM and prompt style used to generate the explanations for each quantum algorithm. Hence, concerning \textbf{RQ$_1$}, \textbf{RQ$_2$}, and \textbf{RQ$_3$}, for each quantum algorithm, the human-raters had to evaluate 18 different explanations (i.e., explanations from 3 LLMs $\times$ 2 prompts styles $\times$ 3 runs) without knowing the LLM and prompt style that generated them. For the \textbf{RQ$_{4}$}, the raters had to evaluate three explanations (one for each LLM) for each quantum algorithm. 

After collecting the evaluations, we performed the following steps to answer each RQ. First, for each explanation, we compute the mean among the scores given by each evaluator. Next, to answer the \textbf{RQ$_1$}, we compare the mean and median scores of all explanations generated from each LLM. In addition, we perform the non-parametric \textit{Wilcoxon signed-ranked} test \cite{woolson2005wilcoxon} to assess if there is a statistically significant difference among the scores of each LLM. We adopt this test because the experiment follows a \textit{one factor with two treatments paired comparison} design (i.e., all multiple subjects evaluate all different treatments) \cite{wohlin_experimentation_2012}. Moreover, we perform this test instead of the parametric \textit{Paired t} test \cite{hsu2014paired} because the data does not follow a normal distribution, as confirmed by the \textit{Shapiro-Wilk} test for normality \cite{razali2011power}. This same evaluation process has also been performed to evaluate LLM improved explanations to assess the \textbf{RQ$_4$}.

To answer the \textbf{RQ$_2$}, we compare the mean and median scores of all explanations generated by each LLM using the \textit{Non Context-aware} and \textit{Context-aware} prompt styles. As done for the \textbf{RQ$_1$}, we employ the \textit{Wilcoxon signed-ranked} test to assess if, for each LLM, there is a statistically significant difference in the scores of explanations obtained using \textit{Context-aware} and \textit{Non context-aware} styles.

To answer the \textbf{RQ$_3$}, we evaluate both the qualitative and syntactical similarity of explanations generated by each LLM over multiple rounds. To assess the qualitative similarity, we compare the mean and median scores of explanations obtained by an LLM with a given prompt style over multiple rounds. We also employ the non-parametric \textit{Kruscal-Wallis H} test \cite{mckight2010kruskal} to assess if there is a statistically significant difference in the scores obtained over multiple rounds. Again, we adopt this test instead of the parametric \textit{ANOVA} \cite{st1989analysis} since the data does not follow a normal distribution. To assess the syntactical similarity, we compute the \textit{cosine similarity} among the explanations generated for the same algorithm over multiple rounds of the same LLM with a specific prompt style. \textit{Cosine similarity} is a widely adopted metric in the Natural Language Processing and Information Retrieval domains, which measures the similarity of two documents as the cosine of their term-frequency vector representation \cite{salton1988term}. This metric ranges from 0 to 1, where 0 means that two documents are entirely different, while 1 means that two documents are syntactically identical. Since we repeated the generation process three times, we report the mean cosine similarity among the three documents. 

Following standard practices \cite{arcuri2011,wohlin_experimentation_2012}, we consider a statistical test significant if its p-value is $<0.05$. However, since we perform multiple hypothesis testing for each RQ (3 tests for \textbf{RQ$_1$}, \textbf{RQ$_2$} and \textbf{RQ$_4$} and 6 tests for \textbf{RQ$_3$}),
we apply the \textit{Bonferroni} correction \cite{weisstein2004bonferroni} and consider a test significant if its p-value is $< 0.05/3$ for \textbf{RQ$_1$}, \textbf{RQ$_2$} and \textbf{RQ$_4$} and $< 0.05/6$ for \textbf{RQ$_3$}. In addition, following the standard guidelines in \cite{wasserstein2016asa}, we support the obtained p-values for \textbf{RQ$_1$}, \textbf{RQ$_2$} and \textbf{RQ$_4$} with the \textit{Cliff's $\delta$} effect size to assess the difference magnitude \cite{macbeth2011cliff}.   

To assess the reliability of human evaluation scores, we use the \textit{Krippendorff's} $\alpha$ inter-rater agreement metric, defined as:
$$
\alpha = \frac{p_a - p_e}{1 - p_e}
$$
where $p_a$ represents the observed weighted percentage agreement (i.e., how often the reviewers actually agreed) and $p_e$ represents the chance weighted percentage agreement (i.e., the percentage agreement the raters would achieve with random scores) \cite{krippendorff2011computing}. This metric ranges from -1 to 1, where -1 indicates a systematic disagreement, 0 means random guessing, and 1 means total agreement among the evaluators. We adopted this metric despite other widely adopted inter-rater agreement metrics like \textit{Cohen's $\kappa$} because it enables the comparison of more than two raters and handles \textit{interval} ratings (i.e., Likert scale evaluations).

\section{Results}\label{sec:results}

In the following, we report the results for our RQs. Concerning the reliability of the human evaluations, the \textit{Krippendorff's $\alpha$} reported an inter-rater agreement of $0.47$ for the evaluation conducted to answer \textbf{RQ$_1$}, \textbf{RQ$_2$}, \textbf{RQ$_3$}, and a result of 0.9 in the evaluation conducted to answer \textbf{RQ$_4$}, highlighting an overall positive agreement among the respondents.  

\subsection{RQ$_1$: LLM-Driven Quantum Explanations}

\begin{table}[tb]
    \centering
   \caption{Mean and median explanation scores for each LLM.}
    \label{tab:rq1}
  \begin{tabular}{l|rrr}
    \toprule
     & \textbf{Gpt3.5} & \textbf{Llama2} & \textbf{Tinyllama} \\
    \midrule
    Mean & 2.50 & \textbf{2.78} & 1.87 \\
    Median & 2.50 & \textbf{2.75} & 1.86 \\
    \bottomrule
    \end{tabular}
\end{table}

\begin{table}[tb]
    \centering
    \caption{Statistic and Bonferroni corrected p-value of the Wilcoxon test and Cliff's $\delta$ among LLM explanation scores.}
    \label{tab:rq1_stats}
    \begin{tabular}{l|rrr}
        \toprule
         & \textbf{W Stat} & \textbf{p-value} & \textbf{$\delta$} \\
        \midrule
        Gpt3.5 - Llama2 & 141.50 & 0.02 & -0.18\\
        Gpt3.5 - Tinyllama & 53.50 & $3.07*10^{-5}$ & 0.44\\
        Llama2 - Tinyllama & 38.00 & $1.01*10^{-6}$ & 0.52 \\
        \bottomrule
    \end{tabular}
\end{table}

Table \ref{tab:rq1} reports the mean and median scores of explanations obtained by each LLM, while Table \ref{tab:rq1_stats} reports the statistics and the Bonferroni corrected p-value of the \textit{Wilcoxon} test and the \textit{Cliff's $\delta$} between each LLM pair. It can be observed from Table \ref{tab:rq1} how Llama2 is the LLM obtaining the highest evaluation score with a mean of 2.78 and a median of 2.75. Gpt3.5 reports instead an average evaluation score of 2.5, which is close to Llama2, as confirmed by a small effect size but still statistically significant, as shown in Table \ref{tab:rq1_stats}. Concerning Tinyllama, we observe instead a lower mean (1.87) and median (1.86) scores compared with both Gpt3.5 and Llama2. This difference is also confirmed by a lower p-value of the \textit{Wilcoxon test} and a greater effect size.  

\begin{figure}[tb]
    \centering
    \includegraphics[width=\linewidth]{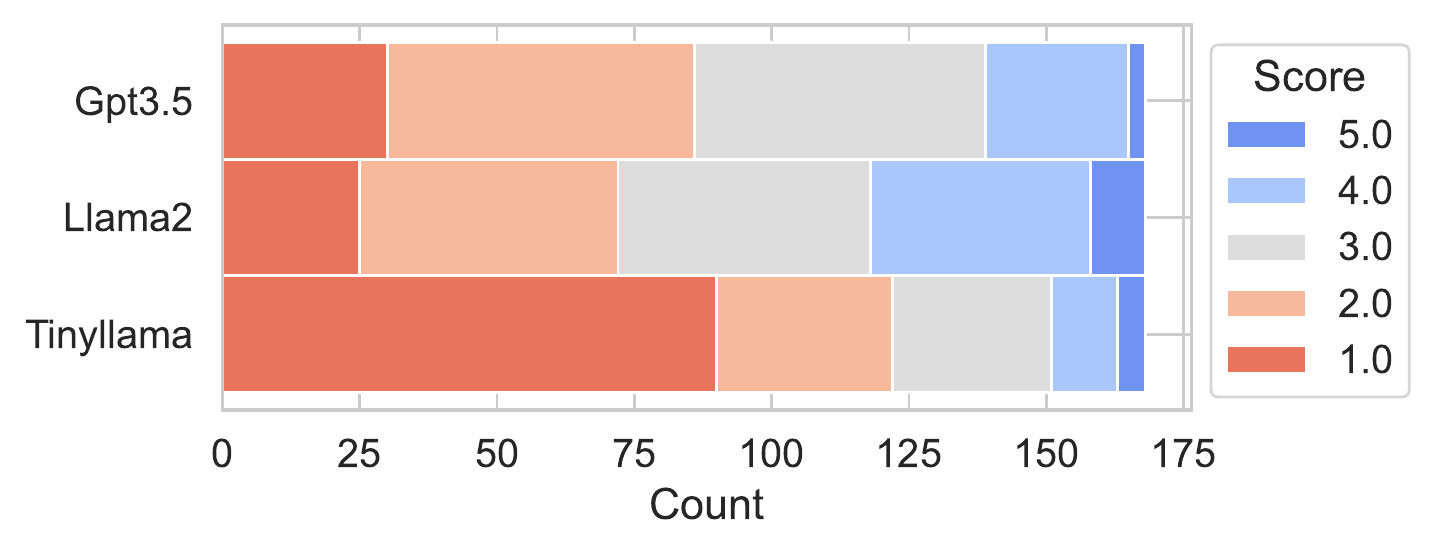}
    \caption{Distribution of explanation scores for each LLM}
    \label{fig:llm_scores}
\end{figure}

Figure \ref{fig:llm_scores} reports the distribution of explanation scores for each LLM. While we observe a quite similar distribution of scores between Gpt3.5 and Llama2, Tinyllama reports a significantly higher amount of 1.0 scores compared with the other LLMs, meaning that many explanations from Tinyllama were systematically wrong.

\textbf{Answer to RQ$_1$:} Llama2 emerged as the LLM that provides better explanations. On the other side, Tinyllama provides many explanations that are systematically wrong compared with the other LLMs.

\subsection{RQ$_2$: Context-Aware Prompts}

\begin{table}[tb]
    \centering
    \caption{Mean and median values, Wilcoxon test and Cliff's $\delta$ between scores of explanations obtained with \textit{Context-aware} and \textit{Non context-aware} prompt styles}
    \label{tab:rq2}
    \resizebox{\linewidth}{!}{\begin{tabular}{l|rr|rr|rrr}
    \toprule
     & \multicolumn{2}{|c|}{\textbf{Non context-aware}} & \multicolumn{2}{|c|}{\textbf{Context-aware}} & \multirow{2}{*}{\textbf{W Stat}} & \multirow{2}{*}{\textbf{p-value}} & \multirow{2}{*}{$\delta$} \\ \cmidrule{2-5}
     & \textbf{Mean} & \textbf{Median} & \textbf{Mean} & \textbf{Median} & & &\\
    \midrule
    Gpt3.5 & 1.86 & 1.75 & \textbf{3.14} & \textbf{3.25} & 0.00 & $2.86*10^{-6}$ & -0.98 \\
    Llama2 & 2.02 & 2.00 & \textbf{3.54} & \textbf{3.75} & 2.00 & $8.58*10^{-6}$ & -0.89 \\
    Tinyllama & 1.36 & 1.25 & \textbf{2.38} & \textbf{2.25} & 1.50 & $3.21*10^{-4}$ & -0.77\\
    \bottomrule
    \end{tabular}}
\end{table}

Table \ref{tab:rq2} reports the mean and median values and the results of \textit{Wilcoxon} test and \textit{Cliff's $\delta$} between the scores of explanations obtained with \textit{Non context-aware} and \textit{Context-aware} prompt styles. From the table, we observe how adding basic information to the prompt (i.e., the name of the algorithm implemented and the number of qubits employed) significantly improves the quality of the explanations in all LLMs. 

\begin{figure}[tb]
    \centering
    \includegraphics[width=\linewidth]{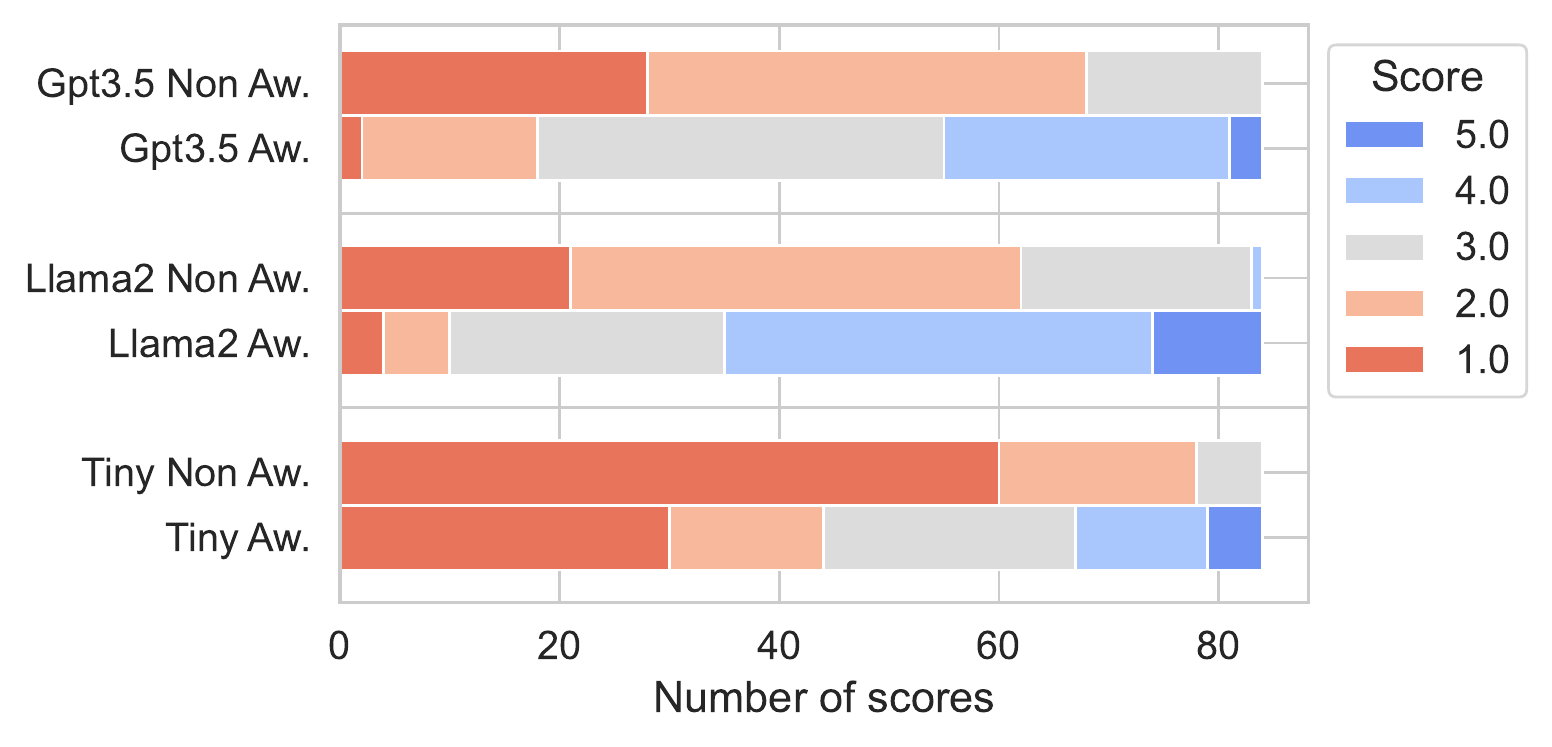}
    \caption{Score distribution for each LLM and prompt style.}
    \label{fig:rq2_distr}
\end{figure}

The higher quality of explanations obtained from a \textit{Context-aware} prompt style is also confirmed by the distribution of scores shown in Figure \ref{fig:rq2_distr}. From the figure, we observe how the proportion of explanations with a low score (i.e., 1 or 2) significantly decreases, especially in Gpt3.5 and Llama2.

\textbf{Answer to RQ$_2$:}
Adding basic information like the name of the algorithm implemented or the number of qubits employed significantly improves the quality of explanations for all the LLMs employed.

\subsection{RQ$_3$: Explanation Consistency}

We assess the similarity of explanations from both a qualitative (i.e., if there is a difference in the quality of explanations) and syntactical (i.e., if the explanations are written differently) point of view.

\begin{table}[tb]
    \centering
    \caption{Mean and median values and Kruscall-Wallis H test result between scores of explanations obtained over three generation rounds for each LLM.}
    \label{tab:rq3}
    \resizebox{\linewidth}{!}{\begin{tabular}{ll|rrrr|rrrr}
    \toprule
     & & \multicolumn{4}{|c|}{\textbf{Non context-aware}} & \multicolumn{4}{|c}{\textbf{Context-aware}} \\\cmidrule{3-10}
     & & \textbf{Mean} & \textbf{Median} & \textbf{H Stat} & \textbf{p-value} & \textbf{Mean} & \textbf{Median} & \textbf{H Stat }& \textbf{p-value} \\
    \midrule
    \multirow{3}{*}{Gpt3.5} & Round 1 & 2.04 & 2.00 & \multirow{3}{*}{3.56} & \multirow{3}{*}{1.01} & 3.32 & 3.50 & \multirow{3}{*}{2.48} & \multirow{3}{*}{1.74} \\
    & Round 2 & 1.71 & 1.75 &  &  & 2.93 & 2.75 &  &  \\
    & Round 3 & 1.82 & 1.75 &  &  & 3.18 & 3.25 &  & \\\midrule
    \multirow{3}{*}{Llama2} & Round 1 & 2.07 & 2.00 & \multirow{3}{*}{1.33} & \multirow{3}{*}{3.08} & 3.43 & 3.75 & \multirow{3}{*}{5.05} & \multirow{3}{*}{0.48} \\
    & Round 2 & 2.07 & 2.00 &  &  & 3.96 & 4.00 &  &  \\
    & Round 3 & 1.93 & 2.00 &  &  & 3.21 & 3.25 &  &  \\\midrule
    \multirow{3}{*}{Tinyllama} & Round 1 & 1.43 & 1.25 & \multirow{3}{*}{1.44} & \multirow{3}{*}{2.91} & 2.36 & 2.25 & \multirow{3}{*}{0.28} & \multirow{3}{*}{5.21} \\
    & Round 2 & 1.39 & 1.25 &  &  & 2.32 & 2.75 &  &  \\
    & Round 3 & 1.25 & 1.00 &  &  & 2.46 & 2.75 &  &  \\
    \bottomrule
    \end{tabular}}
\end{table}

Concerning qualitative similarity, Table \ref{tab:rq3} reports the mean and median scores and the \textit{Kruskall-Wallis H} test result of explanations obtained from three generation rounds of each LLM with a given prompt style. We observe an overall not significant difference in the scores, especially using a \textit{Non context-aware} prompt style. The difference in scores is also emphasised by an overall high p-value of the \textit{H} test, which does not let us reject the null hypothesis of equal medians among the groups.

\begin{figure}[tb]
    \centering
    \includegraphics[width=\linewidth]{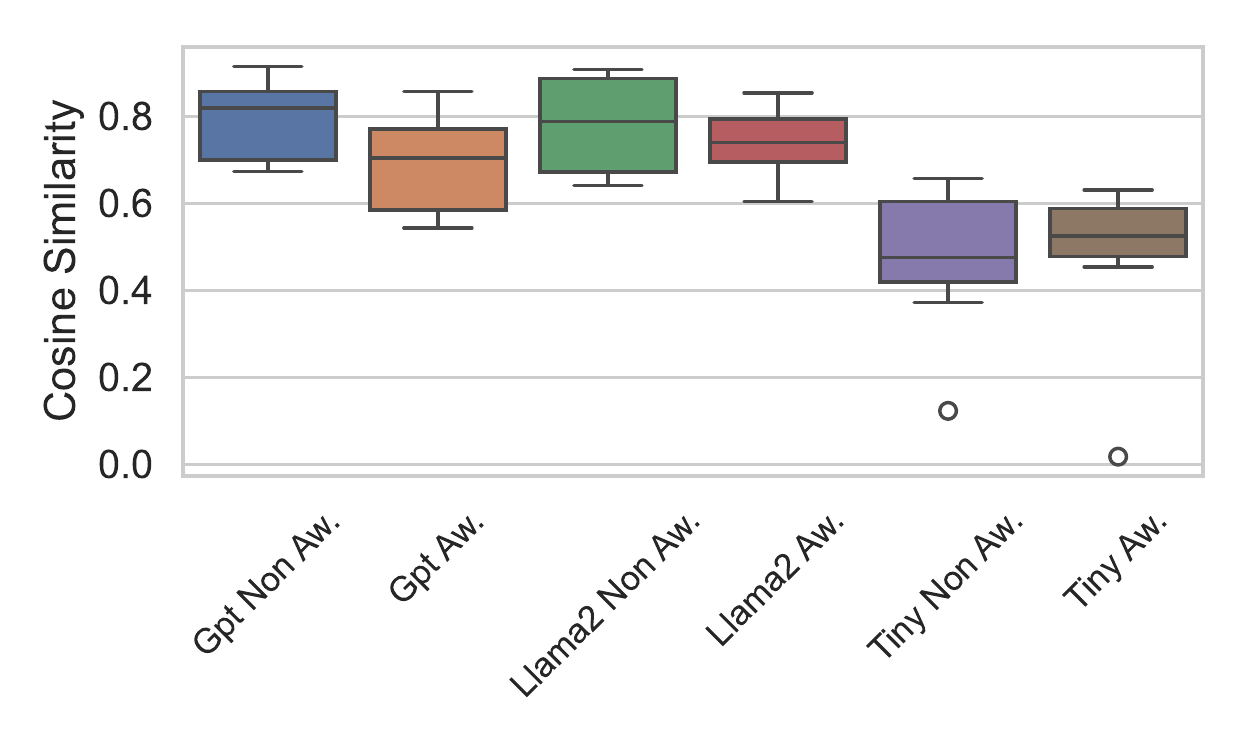}
    \caption{Distribution of cosine similarity of explanations obtained by each LLM with specific prompt style over three generation rounds.}
    \label{fig:cos-sim}
\end{figure}

Concerning syntactic similarity, Figure \ref{fig:cos-sim} reports the distribution of cosine similarity scores among each algorithm's explanation over three generation rounds for each LLM and prompt style pair. We observe how explanations from \textit{Gpt3.5} and \textit{Llama2} have a high similarity score despite the adopted prompt style (with median scores ranging between 0.7 and 0.8). A lower similarity score is instead observed for \textit{Tinyllama}, which can be partially explained by the randomness of some explanations returned by the model. 

\textbf{Answer to RQ$_3$:} Explanations returned by the LLMs over different rounds are overall similar from a qualitative point of view. Concerning the syntactical similarity, \textit{Gpt3.5} and \textit{Llama2} return similar explanations, while more variability is observed in the explanations from \textit{Tinyllama}.

\subsection{RQ$_4$: Quantum Explanation Completion}

\begin{table}[tb]
\centering
\caption{Mean and median scores of improved explanations.}
\label{tab:rq4}
\begin{tabular}{l|rrr}
\toprule
\multicolumn{1}{l|}{} & \multicolumn{1}{l}{\textbf{Gpt3.5}} & \multicolumn{1}{l}{\textbf{Llama2}} & \multicolumn{1}{l}{\textbf{Tinyllama}} \\ \midrule
Mean & \textbf{3.54} & 2.39 & 1.89 \\
Median & \textbf{4.50} & 2.25 & 1.50 \\ \bottomrule
\end{tabular}
\end{table}

Table \ref{tab:rq4} reports the mean and median scores of LLM improved explanations. We notice that Gpt3.5 performs better then the others. However, the results of the \textit{Wilcoxon} test do not allow us to reject the null hypothesis that the distribution of scores is the same for all LLMs, providing p-values $> 0.05/3$ for all comparisons (i.e., 0.375 in the comparison between Gpt3.5 and Llama2, 0.297 between Gpt3.5 and Tinyllama, and 0.247 between Llama2 and Tinyllama). This result may be due to the small dimension of samples (i.e., 7 scores for each LLM) which may lead us to the Type II statistical error (i.e., accept the null hypothesis when it should be rejected) \cite{wohlin_experimentation_2012}.

\textbf{Answer to RQ$_4$:} Gpt3.5 emerges as the LLM performing better in improving existing explanations. However, further research should be conducted to better explore the capabilities of LLMs in improving existing quantum explanations.

\section{Threats to Validity}\label{sec:threats} 

\textbf{Internal Validity:} It is worth noting that our panel of experts possesses some background knowledge in quantum programming, which may appear counter-intuitive to our initial hypothesis of analysing the ability of LLMs to explain quantum code to non-expert software engineers. Nevertheless, we argue that error-free explanations hold greater value. Consequently, in this preliminary evaluation, soliciting evaluations from individuals lacking quantum knowledge might bias results toward superficial aspects, potentially leading to erroneous conclusions. Additionally, all our human-raters boast software engineering backgrounds and have recently transitioned into the realm of quantum computing (see demographic description in Section \ref{sec:method}). Thus, they acknowledge their ongoing learning curve in this domain and recognise the utility of such a tool, notwithstanding their ability to discern basic errors. This criterion may be relaxed in the future once our methodology matures, enabling us to pre-select accurate explanations for a comprehensive user study. In addition, the quality of each explanation has been evaluated using a single question system, which may not be best suited given the multi-faced definitions of \textit{quality}. We plan to extend our analysis with a more structured evaluation system in future works.

\textbf{Construct Validity:}  The quality assessment of each explanation is based on a subjective evaluation, and it may differ among evaluators. To mitigate this threat, we employed the \textit{Krippendorff's} $\alpha$ inter-rater agreement score, which reported an overall positive agreement among the evaluators. 

\textbf{Conclusion Validity:} The results of \textbf{RQ$_4$} have a low statistical significance. This may be due to the small data sample employed in this RQ, which may lead to a Type II statistical error (i.e., refusing to reject the null hypothesis when it should be rejected).  

\textbf{External Validity:} Our analysis focuses on a limited set of LLMs and quantum algorithms, and the results may not hold for other settings. To mitigate this, we consider widely adopted LLMs and state-of-the-art quantum algorithms. In addition, the results of our analysis are limited to algorithms written in the OpenQASM 3 programming language, while the adoption of higher-level programming languages (like Qiskit, Silq, or Q\#) may lead to different results.

\section{Conclusion and Future Directions}\label{sec:conclusion}
This evaluation has provided valuable insights into the capability of LLMs to explain quantum code, highlighting that both GPT3.5 and Llama2 are suitable for this task. Our findings show that including a small amount of context in the prompt significantly enhances the quality of explanations. Additionally, the explanations generated by the LLMs remain qualitatively and syntactically consistent across multiple iterations. However, further work is required to analyse and improve LLMs' ability to explain quantum code comprehensively. None of the LLMs achieved an average high score (i.e., 4 or higher) in generating explanations from scratch without prior context provided in the prompt. While evaluators generally agreed positively on the quality of the explanations, the inter-agreement $\alpha$ score of 0.47 underscores the need for a more systematic approach to evaluating quantum explanations (e.g., following the guidelines outlined in \cite{nauta_anecdotal_2023}). 
Furthermore, despite existing studies on the quality of LLMs' explanations for various aspects of traditional code (see Section~\ref{sec:related}), special attention should be devoted to quantum code. The use of logical qubits and quantum gates is not intuitive for software engineers, and writing algorithms using logical circuits can be cumbersome. This makes quantum code significantly different from classical code and necessitates a unique approach to explanation, as discussed in Section~\ref{sec:background}.

Given these observations, several future research directions emerge. Future work could expand the analysis of GPT3.5 and Llama2 explanations by developing more systematic guidelines for assessing the quality of quantum explanations, involving individuals with varying expertise in both software engineering and quantum computing and employing less-known quantum algorithms. Research should also focus on prompt optimisation for quantum code explanation, determining the minimum amount of context required in the prompt to achieve high-quality explanations. The question of determinism also arises, warranting experiments exploring different temperature settings to observe their impact on the results, especially concerning their consistency over multiple rounds. Additionally, exploring more language models (like GPT4 or Llama3) to investigate their efficiency and efficacy will help identify the best-fitting architecture for this task. Finally, an interesting avenue for further optimisation is response parsing. Implementing a parser to structure the explanations provided by the LLMs can significantly enhance the clarity of these explanations for new users.

\begin{acks}
\footnotesize
G. d'Aloisio is partially funded by the \SoBigDataITAck. S. Fortz is partially supported by the  EPSRC project on Verified Simulation for Large Quantum Systems (VSL-Q), grant reference EP/Y005244/1, the EPSRC project on Robust and Reliable Quantum Computing (RoaRQ), Investigation 009 Model-based monitoring and calibration of quantum computations (ModeMCQ), grant reference EP/W032635/1 and by the QAssure project from Innovate UK. E. Mendiluze Usandizaga is supported by Simula's internal strategic project on quantum software engineering. D. Fortunato is suported by FCT through grant ref.\ 2023.02439.BD. F. Sarro is supported by the ERC Grant no. 741278 and the UCL-CS Strategic Research Fund.
\end{acks}
\normalsize
\bibliographystyle{ACM-Reference-Format}
\bibliography{bibliography}
\end{document}

%% file: commands.tex
\usepackage{graphicx}
\usepackage{xcolor}
\usepackage{tikz,pifont}
\usepackage{amsfonts,amsmath}
\usepackage{hyperref}
\usepackage{xspace}
\usepackage{listings}
\usepackage{wrapfig}
\usepackage{multirow}

\newcommand{\SoBigDataITAck}{European Union - NextGenerationEU - National Recovery and Resilience Plan (PNRR) - Project: “SoBigData.it - Strengthening the Italian RI for Social Mining and Big Data Analytics” - Prot. IR0000013 - Adv. n. 3264 of 28/12/2021\xspace}

\lstdefinestyle{promptstyle}{
	basicstyle=\scriptsize,
	breaklines=true,
	captionpos=b,
	numbers=none,
	numbersep=4pt,
	showspaces=false,
	showstringspaces=false,
	showtabs=false,
	tabsize=2,
	frame=single
}

